%% file: main.tex
\documentclass{article}
\usepackage{spconf,amsmath,graphicx,hyperref}
\usepackage{enumitem}
\usepackage{amsmath, amssymb}
\usepackage{booktabs}
\usepackage{multirow}
\usepackage{subcaption}

\title{Beyond Transport Cost: Routing Differences \\ between Flow Matching and Optimal Transport}
\name{
  Eungyeol Han\qquad
  Jong-Seok Lee \sthanks{Corresponding author}
}
\address{
  \begin{tabular}{c}
    School of Integrated Technology, Yonsei University
  \end{tabular}
}

\begin{document}
\ninept
\maketitle

\input{sec/0_abstract}
\input{sec/1_intro}
\input{sec/2_rework}
\input{sec/3_method}
\input{sec/4_exp}
\input{sec/6_conclusion}

\clearpage

\bibliographystyle{IEEEbib}
\bibliography{strings,refs}

\end{document}

%% file: sec/0_abstract.tex
\begin{abstract}
In generative models, Optimal Transport (OT) is used to improve Flow Matching (FM) by reducing noise-data coupling cost. However, different noise-to-output assignments can yield nearly equal costs, raising a key question. Is cost alone sufficient to guide coupling design? We address this question by separating transport cost from routing, i.e., the destination reached by each noise sample. We show numerically how FM and OT can differ in routing while remaining close in cost. We examine its consequences in learned neural FM. Using the exact FM
routing as an oracle, we further construct a routing-aware training coupling and find that it yields a directionally consistent improvement in generation over a cost-matched, cost-only counterpart. Our findings highlight what cost minimization can overlook and motivate using both cost and routing to evaluate the design of OT-based FM couplings. Code will be released upon acceptance.
\end{abstract}

\begin{keywords}
Flow Matching Models, Optimal Transport
\end{keywords}

%% file: sec/1_intro.tex
\vspace{-0.5em}
\section{Introduction}
\label{sec:Intro}
\vspace{-0.5em}
Flow Matching (FM) generates data such as images by learning a velocity field that transports noise samples to a data distribution. A conventional formulation randomly pairs noise and data samples and trains on straight interpolations between them \cite{liu2022flow, lipman2022flow}. However, these resulting generation trajectories can be curved, making accurate few-step generation difficult \cite{liu2022flow, geng2026mean,pooladian2023multisample, karras2022elucidating}. This has motivated more structured training couplings, which specify how noise samples are paired with data samples.

Optimal Transport (OT) \cite{villani2009optimal, cuturi2013sinkhorn} provides a principled approach to constructing these couplings. It makes trajectories straighter, which enables efficient few-step generation. Quadratic OT \cite{brenier1991polar, gabriel2019computational} minimizes the average squared Euclidean distance between paired samples, and minibatch OT \cite{pooladian2023multisample} approximates this within training batches. That is, by using OT to optimize coupling cost, we make the coupling cost of FM similar to that of OT.

\begin{figure}[t]
    \centering
    \includegraphics[
        width=0.85\columnwidth,
        trim=0pt 0pt 3pt 0pt,
        clip
    ]{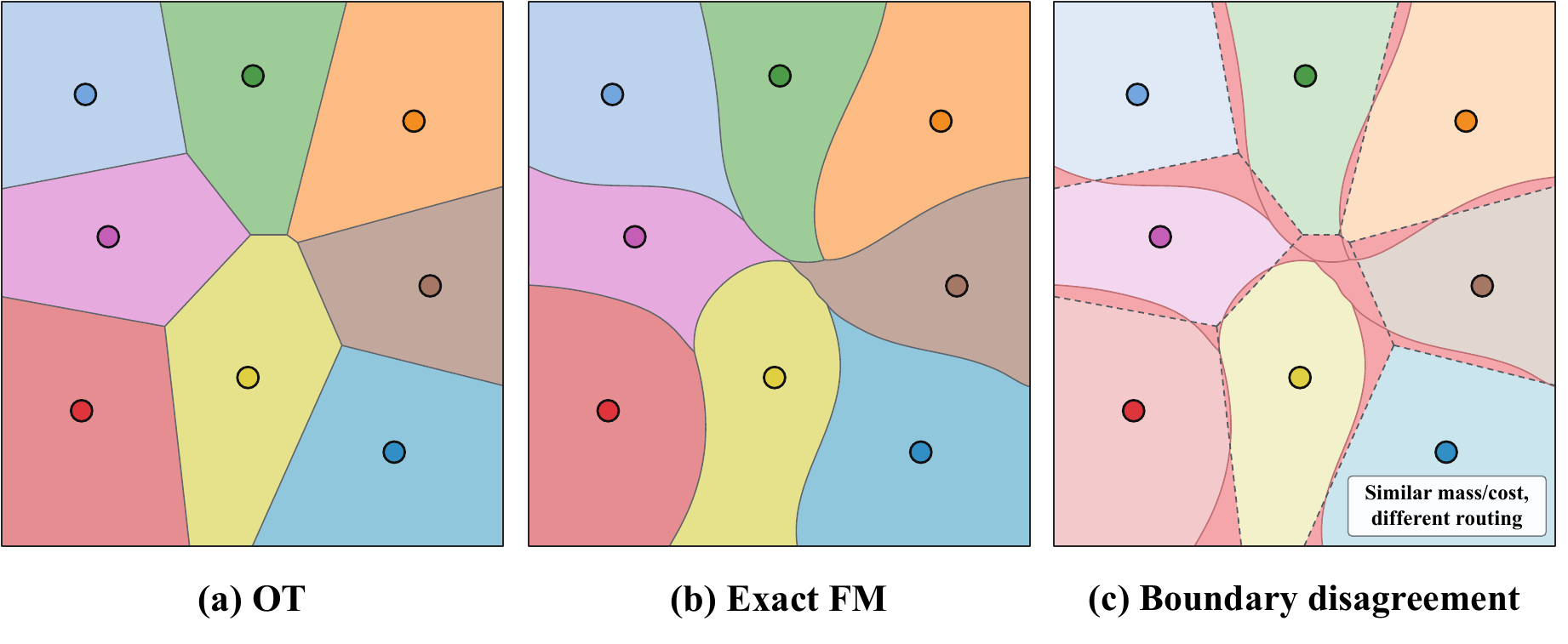}
    \vspace{-2mm}
    \caption{Similar transport cost but different routing between FM and OT.
    Noise samples in each region arrive at the same target atom in the data space.}
    \label{fig:overview}
    \vspace{-5mm}
\end{figure}

However, recent research \cite{hertrich2026relation, pierret2026tessellations} shows that even if FM and OT have the same cost, the routing-- which specifies exactly where they are placed-- may differ. To accurately observe the differences in routing, prior studies adopt a semi-discrete FM setting. In this setting, the source distribution is defined as a continuous distribution (e.g., Gaussian distribution), while the target is defined as discrete atoms. Therefore, a routing and  boundary can be obtained depending on which atoms each initial noise in the source distribution is assigned to (as shown in Figure \ref{fig:overview}).

 This leads to an overlooked question: \textbf{If OT and FM achieve nearly the same transport cost while routing the same noise samples to different destinations, is minimizing coupling cost alone sufficient for FM?}

To answer this question, we investigate the quantitative relationship
between FM and OT beyond simply examining a counterexample. Our
contributions are as follows:

(i) We show that \textbf{similar transport cost does not imply similar routing}, and quantify this gap between FM and OT in a semi-discrete setting. (ii) We empirically validate that this cost--routing gap is observed in \textbf{learned neural FM}. (iii) We demonstrate that a \textbf{routing-aware training objective} yields a directional improvement in generation over a cost-only coupling.

%% file: sec/2_rework.tex
\vspace{-0.8em}
\section{Related Work}
\label{sec:rework}

\vspace{-0.8em}
\textbf{Optimal Transport for Flow Matching.}
FM learns a time-dependent vector field by coupling source and target
samples and regressing the averaged velocity along the interpolation
path \cite{liu2022flow, lipman2022flow, albergo2025stochastic}. OT
improves on random coupling by pairing sources and targets via
shortest-path assignment, reducing coupling cost
\cite{tong2023improving}. In practice this is approximated with
mini-batch OT \cite{fatras2021minibatch,
tong2023improving, pooladian2023multisample}. Couplings are evaluated
mainly through transport cost and downstream metrics such as FID,
leaving routing largely unexamined \cite{damodaran2018deepjdot}.

\vspace{0.2pt}
\noindent\textbf{Geometric Difference between FM and OT.}
FM does not generally recover the OT map. Rectified flow \cite{liu2022flow} reduces cost
through iterative reflow, but the result remains suboptimal
\cite{hertrich2026relation}. Prior work \cite{pierret2026tessellations} offer some counterexamples showing that FM and OT can differ. However, it remains unclear how this
difference in assignments arises numerically and how it affects
generation. This distinguishes our work from neural OT methods that
directly learn the OT map \cite{makkuva2020optimal, rout2021generative}.

%% file: sec/3_method.tex

\vspace{-0.8em}
\section{Proposed Method}
\label{sec:method}
\vspace{-0.8em}

We aim to compare the rates at which cost and routing in FM and OT diverge. By demonstrating that cost diverges more slowly than routing, we highlight the risk of overlooking differences in routing when we optimize only the cost. 

In Section \ref{sec:setup}, we define $D(\theta)$ and $\Delta C(\theta)$ to compare them through precise quantification. $D(\theta)$ represents the difference in routing and $\Delta C(\theta)$ represents the difference in cost. To enable an accurate comparison, we refine the semi-discrete FM setting \cite{pierret2026tessellations} to a 4-target atom setting, as shown in Figure \ref{fig:partitions}. 

In Section \ref{sec:boundary}, we observe how the routing changes each in FM and OT by moving one atom that changes routing. We define $S_\text{FM}$ and $S_\text{OT}$ to measure boundaries where the routing changes and $q(\beta)$ as the difference between them. And we examine $q(\beta)$ in closed form and show how it changes.

In Section \ref{sec:asymmetry}, we go beyond moving one atom to moving the whole boundary. By observing this whole boundary, we measure the difference in the rates at which $D(\theta)$ and $\Delta C(\theta)$ differ in closed form.

Finally, Section \ref{sec:highdim} extends these results in the two-dimensional space to a general ambient dimension $d$.

\vspace{-0.8em}
\subsection{Setup}
\label{sec:setup}

\begin{figure}[t]
\centering
\includegraphics[
    width=0.85\columnwidth,
    trim=0pt 0pt 3pt 0pt,
    clip,
    keepaspectratio
]{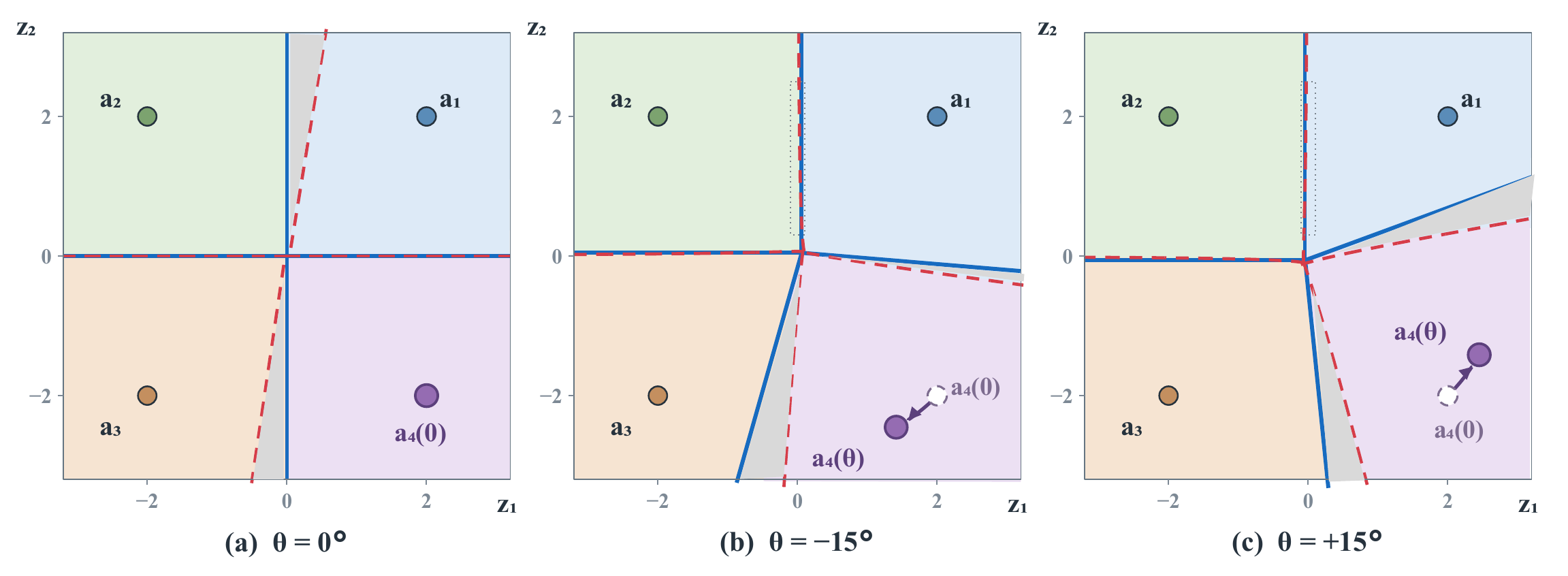}
\vspace{-2mm}
\caption{Different routing-boundary responses of FM and OT to the same atom rotation. Red: Exact FM, Blue: OT.}
\label{fig:partitions}
\vspace{-5mm}
\end{figure}

We use a semi-discrete setting \cite{pierret2026tessellations, le2017existence}: a 2-dimensional continuous Gaussian source
$\mathbf Z\sim \mathcal{N}(\mathbf0,I_2)$ and four target atoms with
law $\nu_\theta=\frac14\sum_{i=1}^4\delta_{\mathbf a_i^\theta}$, where
$\delta_{\mathbf a}$ is a unit point mass and
\begin{equation}
\begin{aligned}
\mathbf a_1^\theta&=r(1,1),
&\mathbf a_2^\theta&=r(-1,1),\\
\mathbf a_3^\theta&=r(-1,-1),
&\mathbf a_4^\theta&=rR_\theta(1,-1).
\end{aligned}
\label{eq:target-family}
\end{equation}
Here $r>0$ is fixed, $R_\theta$ is counterclockwise rotation by $\theta$
radians, and $|\theta|<\pi/4$. As shown in Figure \ref{fig:partitions}, starting from $\theta=0$ where both cost and routing are identical in FM and OT we will vary $\theta$ to move the atom and observe how the routing and boundary change.

We can obtain the exact velocity of FM in this setting, which is used to compute where FM actually sends each noise sample, and to compute the cost of the map. For independent $\mathbf Z$ and $\mathbf X_1\sim\nu_\theta$, we set
$\mathbf X_t=(1-t)\mathbf Z+t\mathbf X_1$, where $t\in[0,1]$ by linear interpolation \cite{liu2022flow,lipman2022flow,pierret2026tessellations}.  Before calculating the velocity, we compute the posterior probability that
$\mathbf{X}_i=\mathbf{a}_i^\theta$ given $\mathbf{X}_t=\mathbf{x}$. This probability follows directly from Bayes' rule since each atom gives a Gaussian distribution over $\mathbf{x}$, 
\begin{equation}
p_i^\theta(t,\mathbf x)
=\frac{e^{-\|\mathbf x-t\mathbf a_i^\theta\|^2/[2(1-t)^2]}}
{\sum_j e^{-\|\mathbf x-t\mathbf a_j^\theta\|^2/[2(1-t)^2]}}.
\end{equation} 

Then, using $p_i^\theta(t,\mathbf{x})$, the velocity $v_t^\theta(\mathbf{x})$ is obtained as
\begin{equation}
\label{eq:velocity}
v_t^\theta(\mathbf x)
=\frac{\sum_i p_i^\theta(t,\mathbf x)\mathbf a_i^\theta-\mathbf x}{1-t}.
\end{equation}

We define $D(\theta)$ and $\Delta C(\theta)$ to evaluate differences in routing and cost between FM and , respectively. 
\begin{equation}
\begin{aligned}
D(\theta)&=\Pr[k_{\mathrm{FM}}^\theta(\mathbf Z)
                 \ne k_{\mathrm{OT}}^\theta(\mathbf Z)],\\
\Delta C(\theta)&=C(T_{\mathrm{FM}}^\theta)-C(T_{\mathrm{OT}}^\theta)\geq0,
\end{aligned}
\label{eq:routing-cost}
\end{equation}
where $k_M^\theta$ is the destination index for $M\in\{\mathrm{FM},\mathrm{OT}\}$ and $T_M^\theta$ denotes the atom reached by solving the ODE from $\mathbf{z}$. $D(\theta)$ is the probability that FM and OT disagree on which atom to send the noise to. $\Delta C(\theta)$ measures how much extra cost FM pays compared to OT. Since OT achieves the minimum cost, $\Delta C(\theta)\geq0$ always.

\vspace{-0.8em}
\subsection{Boundary response to a single-atom moving}
\label{sec:boundary}

In this section, we show how the exact forms of routing and boundaries differ between FM and OT. Since routing ultimately depends on which atom noise moves to, we will move one of the four atoms set up in Section \ref{sec:setup} by $\theta$ to observe how routing becomes different between FM and OT (see Figure \ref{fig:partitions}). Furthermore, to examine the changes in the boundary where the routing changes, we define $S_\text{FM}$ and $S_\text{OT}$ as the boundaries for FM and OT, respectively, and derive the difference between them  in closed form.

To observe changes in the boundary, we define the horizontal coordinate $s_M(\beta; \theta)$, $M\in\{{\text{FM, OT}}\}$, perpendicular to the boundary and use it as a reference to calculate how the boundaries change. Here $\beta$ refers to height of boundary. Since our goal is to show that the boundary changes rapidly relative to the cost, we take a first-order Taylor expansion of $s_M(\beta;\theta)$ around $\theta=0$:
\begin{equation}
s_M(\beta; \theta)=\theta S_M(\beta)+o(\theta),
\label{eq:boundary-param}
\end{equation}
where $S_M(\beta)$ indicates the linear rate at which $s_M(\beta, \theta)$ moves by $\theta$. $o(\theta)$ accounts for remaining higher-order terms; which vanishes faster than $\theta$.

\noindent\textbf{FM response.}
FM trajectory is denoted as $\dot{\boldsymbol\gamma}_t^\theta=v_t^\theta(\boldsymbol\gamma_t^\theta)$,
starting from a noise point at $t=0$. To track how this trajectory
moves as $\theta$ turns on, we rewrite it in coordinates that remove
$t$ from the equations.

We make three substitutions: rescale the atoms by $r$,
$\mathbf b_i^\theta=\mathbf a_i^\theta/r$, replace $t$ with
$u=rt/(1-t)$, so $u$ runs from $0$ to $\infty$ as $t$ runs from $0$ to
$1$, with the final FM output at $u\to\infty$ and rescale the
trajectory, $\mathbf y=\boldsymbol\gamma_t^\theta/(1-t)$, to remove the contraction term in the velocity field. The exact velocity \eqref{eq:velocity} then
simplifies to
\begin{equation}
\mathbf y'
=\frac{\sum_i\mathbf b_i^\theta e^{u\mathbf y^\top\mathbf b_i^\theta}}
       {\sum_j e^{u\mathbf y^\top\mathbf b_j^\theta}},
\label{eq:rescaled-flow}
\end{equation}
with primes now denoting $u$-derivatives.

At $\theta=0$, the symmetric boundaries decouple the two coordinates of
$\mathbf y$: each $y_j$ evolves independently by $y_j'=\tanh(uy_j)$.
A point started on the boundary between atoms $1$ and $2$ has
$y_1=0$, for all $u$, while its vertical coordinate
evolves as $Y_\beta(u)$, the solution of $y_2'=\tanh(uy_2)$ started at
height $\beta$. We write this boundary trajectory as $(0,Y_\beta(u))$.

We now perturb $\theta$ and ask: how must the starting horizontal coordinate change at fixed $\beta$ to stay on the boundary? A point is on the boundary when it is
equally likely, under FM, to reach either atom-- the
\emph{equal-posterior condition}. Since this cannot be evaluated at
$u=\infty$ directly, we first impose it at a finite time $u=H$, and
let $H\to\infty$ only afterward. Let $s_{\mathrm{FM},H}$ be the
starting coordinate satisfying this finite-time condition.
Differentiating it with respect to $\theta$ at $0$ gives a linear ODE
for the sensitivity $\xi_{\beta,H}$:
\begin{equation}
\begin{gathered}
Y_\beta'=\tanh(uY_\beta),\qquad Y_\beta(0)=\beta,\\
\xi_{\beta,H}'=u\xi_{\beta,H}+g_\beta(u),\quad\xi_{\beta,H}(H)=0,\\
g_\beta(u)=\tfrac14[1-\tanh(uY_\beta)](1+uY_\beta).
\end{gathered}
\label{eq:boundary-sensitivity}
\end{equation}
Here $\xi_{\beta,H}(0)=\partial_\theta s_{\mathrm{FM},H}|_0$ is the
quantity we want, up to horizon $H$. Solving this linear ODE with an
integrating factor gives
\begin{equation}
\label{eq:terminal_response}
\partial_\theta s_{\mathrm{FM},H}|_0=-\int_0^H e^{-u^2/2}g_\beta(u)\,du.
\end{equation}

To recover the terminal response, we let $H\to\infty$, which requires
an assumption:

\emph{Assumption A1.} The finite-horizon boundaries converge, together
with their first derivatives in $(\beta,\theta)$, to their terminal
boundaries as $H\to\infty$, locally near $\theta=0$ and away from the
origin-- i.e., the limit $H\to\infty$ and the derivative in $\theta$
can be exchanged.

A1 lets us take $H\to\infty$ directly inside the integral above,
recovering the terminal response in Eq. \ref{eq:terminal_response}
\begin{equation}
S_{\mathrm{FM}}(\beta)
=-\int_0^\infty e^{-u^2/2}g_\beta(u)\,du,
\label{eq:fm-boundary-response}
\end{equation}
which is always finite since $0\leq g_\beta\leq1/4$.

\noindent\textbf{OT response and signed gap.}
We now compute $S_{\mathrm{OT}}(\beta)$ the same way, but using OT's
rule for assignment instead of an ODE. OT does not simply send each
point to its nearest atom. It adjusts each atom's cost by an offset
$\psi_i^\theta$, called a \emph{dual potential}, and sends each point
to the atom with the lowest adjusted cost,
\begin{equation}
\label{eq:adjusted_cost}
\ell_i^\theta(\mathbf z)=\tfrac12\|\mathbf z-\mathbf a_i^\theta\|^2-\psi_i^\theta.
\end{equation}
These offsets exist because without them, a plain
nearest-atom rule would not send exactly probability $1/4$ of the
Gaussian mass to each atom once the square is perturbed. The dual
potentials $\psi_i^\theta$ are chosen exactly so that this equal-mass
condition holds for all four atoms
\cite{gabriel2019computational,jimenez2012optimal}. We fix the one
free constant in their definition by requiring
$\sum_i\psi_i^\theta=0$.

Because $\ell_i^\theta$ is quadratic in $\mathbf z$, the boundary where
two adjusted costs are equal is always a straight line, unlike the
curved FM boundary above. In particular, the boundary between atoms
$1$ and $2$ sits at $s_{\mathrm{OT}}=(\psi_2^\theta-\psi_1^\theta)/(2r)$ since $\ell_1^\theta(\mathbf{z})-\ell_2^\theta(\mathbf{z})
= -2r\mathbf{z}_1-\psi_1^\theta+\psi_2^\theta
= 0$, where $\mathbf{z}=(\mathbf{z}_1, \mathbf{z_2})$.
This is independent of $\beta$, so the upper OT boundary remains vertical and shifts horizontally. Differentiating the four equal-mass constraints
with respect to $\theta$ gives
\begin{equation}
\begin{gathered}
\left.\partial_\theta \psi_1^\theta\right|_{\theta=0}
= r\varphi_0,
\quad
\left.\partial_\theta \psi_3^\theta\right|_{\theta=0}
= -r\varphi_0,
\\
\left.\partial_\theta \psi_2^\theta\right|_{\theta=0}
=
\left.\partial_\theta \psi_4^\theta\right|_{\theta=0}
= 0,
\quad S_{\mathrm{OT}}(\beta)=-\kappa,
\quad
\kappa=\varphi_0/2,
\end{gathered}
\end{equation}
where $\varphi(\beta)=(2\pi)^{-1/2}e^{-\beta^2/2}$ is the standard
Gaussian density and $\varphi_0=\varphi(0)=1/\sqrt{2\pi}$.

We can now compare the two responses directly. Define their
difference,
\begin{equation}
q(\beta):=S_{\mathrm{FM}}(\beta)-S_{\mathrm{OT}}(\beta)
=\kappa-\int_0^\infty e^{-u^2/2}g_\beta(u)\,du,
\label{eq:analytic-gap}
\end{equation}
so that $s_{\mathrm{FM}}-s_{\mathrm{OT}}=\theta q(\beta)+o(\theta)$.
This tells us, at each height $\beta$, how differently
FM and OT respond to the same small change in $\theta$.

\emph{$q(\beta)$ changes sign.} As $\beta$ increases, $Y_\beta$ grows too
(since $Y_\beta(0)=\beta$ and $Y_\beta'=\tanh(uY_\beta)>0$). This
pushes $g_\beta(u)$ toward $0$ for every fixed $u$, so the integral in
Eq. \ref{eq:analytic-gap} shrinks and $q(\beta)$ increases with $\beta$.
At the two extremes,
\begin{equation}
q(0^+)=\kappa-\tfrac14\sqrt{\pi/2}<0,
\qquad q(\infty)=\kappa>0.
\label{eq:gap-limits}
\end{equation}
Since $q(\beta)$ is monotonic and changes sign, there is exactly one
height $\beta^*$ where $q(\beta^*)=0$. Below
$\beta^*$ and above $\beta^*$, FM and OT disagree in opposite ways:
for small positive $\theta$ below $\beta^*$, FM sends the strip near
the boundary to the upper-left atom while OT sends it to the
upper-right, above $\beta^*$, this reverses. Either way, the two upper
atoms only exchange samples between each other, neither atom's total
share of samples changes.

The same pattern holds at all four boundaries. So far we
studied only the upper boundary. The same steps apply to the other
three boundaries (left, bottom, right) in Figure \ref{fig:partitions}, each measured by its own
axial distance $\beta>0$:
\vspace{-0.5em}
\begin{equation}
\begin{array}{c|c|c}
e&S_{\mathrm{FM}}^e&S_{\mathrm{OT}}^e\\ \hline
\mathrm U (\text{Upper}),\mathrm L(\text{Left})&S_{\mathrm{FM}}&-\kappa\\
\mathrm B (\text{Bottom}),\mathrm R (\text{Right})&S_{\mathrm{FM}}+\beta/2&-\kappa+\beta/2
\end{array}
\label{eq:four-interface-responses}
\end{equation}
\vspace{-0.2em}
with $S_M^e=\partial_\theta s_M^e|_0$. Every boundary shares the same
signed gap $q(\beta)$, even though each one's individual shift
$S_M^e$ differs. In other words, the sign change we found above is
not special to the upper boundary. It appears at all four boundaries
around the square.

\vspace{-0.8em}
\subsection{First-order routing, second-order excess cost}
\label{sec:asymmetry}

Note that $q(\beta)$ is the difference
between the FM and OT boundary response, at each height $\beta$ on one
boundary. This is a \emph{local} quantity. But $D(\theta)$ and
$\Delta C(\theta)$, defined in Eq.
\ref{eq:routing-cost}, are \emph{global}: they average over the
whole noise distribution. In this section, we add up our local result
over every height and every boundary, to obtain these global
quantities. We will see that they grow at different rates as
$\theta\to0$: routing disagreement $D(\theta)$ grows quickly, in
proportion to $\theta$, excess cost $\Delta C(\theta)$ grows
only slowly, in proportion to $\theta^2$. We call these two rates
\emph{first-order} and \emph{second-order} in $\theta$, respectively.

\noindent\textbf{Why routing disagreement is first-order.}
Recall from Section~\ref{sec:boundary} that, at height $\beta$, FM and
OT disagree over a thin strip of width as
\begin{equation}
\label{eq:h}
|h_e|=|s_{\mathrm{FM}}^e-s_{\mathrm{OT}}^e|=|\theta q(\beta)+o(\theta)|.
\end{equation}
This width is proportional to $\theta$. Near the boundary, the density of noise samples does not depend on $\theta$, only the strip's width does. So the probability mass inside the strip is proportional to the width and $\theta$. Since $D(\theta)$ is exactly this mass, summed over the
four strips, $D(\theta)$ grows linearly in $|\theta|$.

\noindent\textbf{{Why excess cost is second-order.}}
Exactly on the OT boundary, sending a point
to either neighboring atom costs the same. Moving away from the
boundary, this equality breaks, and the gap between the two costs grows.
On the upper boundary, this gap grows linearly with the distance
$n$ from the boundary: it equals exactly $2r|n|$. Integrating this
growing penalty across the strip, from the boundary ($n=0$) out to its
far edge ($n=|h_{\mathrm U}|$), gives
\begin{equation}
\label{eq:penalty}
\int_0^{|h_{\mathrm U}|}2rn\,dn=rh_{\mathrm U}^2.
\end{equation}
Since $h_{\mathrm U}$ is $O(\theta)$ as in Eq.\ref{eq:h}, the above is $O(\theta^2)$. This is because \emph{both} the strip's width and the
penalty inside it shrink with $\theta$, so the cost shrinks twice
over, as $\theta^2$ rather than $\theta$. The other three boundaries behave the same way.

Using the reduced costs $\ell_i^\theta=\frac12\|\mathbf z-\mathbf a_i^\theta\|^2-\psi_i^\theta$
from Section~\ref{sec:boundary}, the true cost of atom $k$ is
$\ell_k^\theta+\psi_k^\theta$. FM and OT both send probability $1/4$
to each atom, so the expected offset $\psi_{k_M}^\theta$ is the same,
$\frac14\sum_i\psi_i^\theta$, for both, and cancels in their
difference as
\begin{equation}
\Delta C(\theta)=\mathbb E[
\ell_{k_{\mathrm{FM}}}^\theta(\mathbf Z)
-\ell_{k_{\mathrm{OT}}}^\theta(\mathbf Z)].
\label{eq:dual-cancellation}
\end{equation}

\noindent\textbf{Two more assumptions.}
To add up the four strips rigorously, we need two assumptions. Near
$\theta=0$, all four boundaries meet at the origin, so this is where
their interactions are hardest to control, so we isolate this region in
a small square $J_\theta$ around the
origin, whose size shrinks in proportion to $\theta$.
\emph{A2:} outside $J_\theta$, the disagreement consists of four
disjoint adjacent-atom strips, parameterized by $\beta$. Anything unusual,
such as points where three or more atoms meet, stays confined
inside $J_\theta$.
\emph{A3:} farther out, $|s_M^e(\beta;\theta)|/|\theta|\leq W(\beta)$
for some fixed $W$ whose integral converges. Like A1, these are
hypotheses on the exact terminal boundaries.

\noindent\textbf{{Conditional separation result.}}
Under A1--A3, summing the mass and cost of the four strips, weighted
by the Gaussian density, gives
\begin{equation}
\begin{aligned}
D(\theta)&=A|\theta|+o(|\theta|),\quad \Delta C(\theta)=B\theta^2+o(\theta^2),\\
A&=4\varphi_0\int_0^\infty\varphi(\beta)|q(\beta)|\,d\beta,\\
B&=4r\varphi_0\int_0^\infty\varphi(\beta)q(\beta)^2\,d\beta,
\quad A,B>0.
\end{aligned}
\label{eq:separation-laws}
\end{equation}
The central junction $J_\theta$ contributes only $O(\theta^2)$ mass at
$O(|\theta|)$ penalty each, i.e.\ $O(|\theta|^3)$ total cost, which is negligible next to the $O(\theta^2)$ above.

Equivalently, since $D(\theta)\approx A|\theta|$,
\begin{equation*}
\Delta C(\theta)=\frac{B}{A^2}D(\theta)^2+o(D(\theta)^2):
\end{equation*}
excess cost grows only as the \emph{square} of routing disagreement.
A large routing gap can therefore coexist with a small cost gap.

\vspace{-0.8em}
\subsection{Extension to higher ambient dimensions}
\label{sec:highdim}

For $U\in\mathbb R^{d\times2}$ with $U^\top U=I_2$, we embed the atoms
as $U\mathbf a_i^\theta$ and use source $\mathcal N(\mathbf0,I_d)$.
Orthogonal decomposition reduces both assignments to $U^\top\mathbf z$
\cite{pierret2026tessellations}. Indexing the same metrics
by ambient dimension gives
\begin{equation}
\begin{gathered}
D_d(\theta)=D_2(\theta),\quad\Delta C_d(\theta)=\Delta C_2(\theta),\\
C_{M,d}(\theta)=C_{M,2}(\theta)+(d-2)/2.
\end{gathered}
\label{eq:dimension-reduction-laws}
\end{equation}
Thus $q,A,B$ are unchanged for these planar embeddings.

%% file: sec/4_exp.tex
\vspace{-0.8em}
\section{Experiments}
\label{sec:exp}

We test our theory in three steps. (i) We check whether $q(\beta)$
predicts the boundary shift and the separation law in Eq. \ref{eq:separation-laws} in the exact, synthetic setting. (ii) We
test whether this cost--routing gap survives when FM is learned by a
network, and whether it affects generation quality. (iii) We test
whether the gap extends to image data.

\subsection{Setup}
\label{sec:exp-setup}

\textbf{Exact setting.} We use the Gaussian source and one-target
rotation family of \eqref{eq:target-family}, sweeping $\theta$
(positive, negative, and the symmetric $\theta=0$ control). $q$, $A$,
$B$ are computed once from the unperturbed dynamics and never refit,
so all comparisons below are out-of-sample. FM is integrated to
$t=0.999$. OT offsets are solved independently.

\textbf{Learned FM.} We fix the source bank, targets, network, and
training budget, and vary only the training coupling. We use 3-layer MLP with SiLU
activations and train it with Adam ($\text{lr}=10^{-3}$) for 8{,}000 updates with batch size 512, across ten random seeds. \textbf{Cost Only (CO)}
minimizes pairing cost. \textbf{Routing Aware (RA)} maximizes agreement with the exact
FM routing $k_{\mathrm{FM}}^\theta$ under a cost budget matched to CO
(using exact routing as an oracle). \textbf{Routing Bad (RB)} matches the same cost
budget but maximizes \emph{disagreement} with $k_{\mathrm{FM}}^\theta$
instead.

\vspace{-0.8em}
\subsection{Results}
\label{sec:exp-results}

\begin{table}[t]
\centering
\caption{Numerical validation of the boundary-displacement prediction
and the resulting cost--routing separation in Eq. \ref{eq:separation-laws}.}
\label{tab:exact-validation}
\footnotesize
\setlength{\tabcolsep}{3pt}
\begin{tabular}{@{}llcc@{}}
\toprule
Quantity & Metric & Theory & Measured \\
\midrule
\multirow{3}{*}{Displacement}
& $R^2$, full grid & -- & 0.9986 \\
& $R^2$, $|\theta|\le2^\circ$ & -- & 0.9991 \\
& Sign change at $\beta^*$ & 0.6455 & 0.6455 \\
\midrule
\multirow{2}{*}{Routing mass}
& Log--log slope & 1 & 0.9992 \\
& $A$ vs.\ $D/|\theta|$ & 0.038038 & 0.038040 \\
\midrule
\multirow{2}{*}{Excess cost}
& Log--log slope & 2 & 1.9976 \\
& $B$ vs.\ $\Delta C/\theta^2$ & 0.010222 & 0.010226 \\
\bottomrule
\end{tabular}
\end{table}

\textbf{Table~\ref{tab:exact-validation} confirms the theory in the
exact setting.} The Displacement rows show that our
closed-form prediction $\theta q(\beta)$ matches the measured
boundary shift, including the predicted sign change
at $\beta^*$. The Routing mass and Excess cost rows
show that integrating this local prediction over the whole boundary
reproduces the separation law: log--log slopes of $0.999$ and
$1.998$ confirm that disagreement mass grows linearly in $|\theta|$
while excess cost grows only quadratically, exactly as
Eq. \ref{eq:separation-laws} predicts.

\textbf{Table~\ref{tab:learned-diagnostic} shows that this separation
survives learning, and it matters.} Despite nearly identical
$C$, CO, RA, and RB differ sharply in $D_{\mathrm{FM}}$: matching
cost does not fix routing. This gap carries over to the trained
network, with $D_{\mathrm{net,FM}}$ and $D_{\mathrm{net,OT}}$ lowest
for RA and highest for RB. It also matters for generation. RA
improves both MMD and FD over CO, while RB does not. This improvement is significant ($p<0.05$ from paired t-tests over seeds).

\vspace{-0.8em}
\subsection{Generalization to image data}
\label{sec:mnist}
\vspace{-0.5em}

To test whether this gap extends beyond the synthetic setting, we
repeat the CO/RA/RB comparison on $28\times28$ MNIST ($d=784$) under a
continuous Gaussian source and a ten-class discrete target. We
parameterize the velocity field with a time-conditioned U-Net with Adam optimizer, $\text{lr}=2\times10^{-4}$, and 8,000 updates with batch size 64. Since
exact FM routing is unavailable here, we use the terminal assignments
of a pretrained CO network as the routing oracle for RA and RB under
the same cost constraint as
Section~\ref{sec:exp-setup}.

Table~\ref{tab:mnist} shows the same pattern as before. RA improves generation quality over
CO. This suggests the cost--routing separation is not an artifact of
the four-atom, though large-scale validation remains
future work.

\vspace{-0.8em}
\begin{table}[t]
\centering
\begin{minipage}[t]{0.56\linewidth}
\centering
\caption{Cost-matched learned-FM diagnostic. Coupling metrics are
computed once. Routing and generation metrics report mean $\pm$ std
over ten runs. MMD = Maximum Mean Discrepancy. FD = Fréchet Distance}
\label{tab:learned-diagnostic}
\resizebox{\linewidth}{!}{%
\begin{tabular}{@{}lcccc@{}}
\toprule
 & $C$ & $D_{\mathrm{FM}}\downarrow$
 & $D_{\mathrm{net,FM}}\downarrow$ & $D_{\mathrm{net,OT}}\downarrow$ \\
\midrule
CO & .814 & .020 & $.022\pm.008$ & $.022\pm.008$ \\
RA & .818 & \textbf{.013} & $\mathbf{.020\pm.008}$ & $\mathbf{.020\pm.008}$ \\
RB & .819 & .063 & $.025\pm.008$ & $.025\pm.008$ \\
\bottomrule
\end{tabular}}

\resizebox{\linewidth}{!}{%
\begin{tabular}{@{}lcccc@{}}
\toprule
 & MMD$_1\downarrow$ & MMD$_4\downarrow$ & FD$_1\downarrow$ & FD$_4\downarrow$ \\
\midrule
CO & $.076\pm.011$ & $.038\pm.008$ & $.012\pm.001$ & $.005\pm.002$ \\
RA & $\mathbf{.074\pm.012}$ & $\mathbf{.038\pm.008}$
   & $\mathbf{.009\pm.000}$ & $\mathbf{.003\pm.000}$ \\
RB & $.076\pm.012$ & $.040\pm.006$ & $.011\pm.001$ & $.007\pm.002$ \\
\bottomrule
\end{tabular}}
\end{minipage}
\hfill
\vspace{-1.5em}
\begin{minipage}[t]{0.41\linewidth}
\centering
\caption{Cost--routing diagnostics and generation quality on
$28\times28$ MNIST.}
\label{tab:mnist}
\resizebox{\linewidth}{!}{%
\begin{tabular}{lccc}
\toprule
Coupling & $C$ & $D_{net}\downarrow$ & $\mathrm{FD}\downarrow$ \\
\midrule
CO & 0.818 & 0.025 & 24.72 \\
RA & 0.820 & \textbf{0.011} & \textbf{21.32} \\
RB & 0.821 & 0.071 & 28.54 \\
\bottomrule
\end{tabular}}
\end{minipage}
\end{table}

%% file: sec/6_conclusion.tex
\section{Conclusion}
\label{sec:conclusion}

We showed that FM and OT couplings can attain nearly identical transport cost while routing samples to substantially different destinations and validated these results in learned neural FM. Our findings suggest that cost minimization alone may not fully characterize the quality of an OT-based coupling for FM, motivating routing-aware objectives as a complementary criterion in broader OT-FM
training algorithms.